\documentclass[conference, a4paper]{IEEEtran}
\IEEEoverridecommandlockouts
\usepackage{cite}
\usepackage{amsmath,amssymb,amsfonts}
\usepackage{algorithmic}
\usepackage{graphicx}
\usepackage{textcomp}
\usepackage{xcolor}
\usepackage{stfloats} 
\usepackage{multirow}
\usepackage[OT1]{fontenc}
\usepackage{algorithmic}
\usepackage{algorithm}
\usepackage{setspace}

\def\BibTeX{{\rm B\kern-.05em{\sc i\kern-.025em b}\kern-.08em
    T\kern-.1667em\lower.7ex\hbox{E}\kern-.125emX}}

\makeatletter
\newcommand{\linebreakand}{%
  \end{@IEEEauthorhalign}
  \hfill\mbox{}\par
  \mbox{}\hfill\begin{@IEEEauthorhalign}
}
\makeatother

\renewcommand\IEEEkeywordsname{Keywords}

\begin{document}

\title{Refining Coded Image in Human Vision Layer Using CNN-Based Post-Processing}

\author{\IEEEauthorblockN{Takahiro Shindo, Yui Tatsumi, Taiju Watanabe, Hiroshi Watanabe}
\IEEEauthorblockA{\textit{Waseda University,} Tokyo, Japan}

\vspace{-18pt}
}
\maketitle
\begin{abstract}
\boldmath
Scalable image coding for both humans and machines is a technique that has gained a lot of attention recently.
This technology enables the hierarchical decoding of images for human vision and image recognition models.
It is a highly effective method when images need to serve both purposes.
However, no research has yet incorporated the post-processing commonly used in popular image compression schemes into scalable image coding method for humans and machines.
In this paper, we propose a method to enhance the quality of decoded images for humans by integrating post-processing into scalable coding scheme.
Experimental results show that the post-processing improves compression performance.
Furthermore, the effectiveness of the proposed method is validated through comparisons with traditional methods.
\end{abstract}

\begin{IEEEkeywords}
scalable image coding, image coding for machines, post-processing, image compression
\end{IEEEkeywords}

\section{Introduction}
Rapid advancements in image recognition models are making our lives easier.
These models can estimate the size, location, and type of objects and backgrounds in an image.
However, image analysis using these models is not flawless and often needs human verification.
For example, in traffic monitoring and farm animal management, machines perform the initial image analysis, but human review is still necessary.
Scalable image coding for both humans and machines is a promising technology for these applications.

In traditional image codecs like HEVC\cite{a1} and VVC\cite{a2}, post-processing is an effective way to enhance the quality of decoded images. 
Many studies have also explored using neural networks for post-processing to eliminate coding noise \cite{a3,a4}.
However, no research has yet focused on applying post-processing to scalable coding for humans and machines.
Scalable coding methods decode images for humans by adding information to those decoded for image recognition models, leading to different coding noise than standard image compression methods.
In this paper, we introduce a post-processing to reduce this coding noise.
Experimental results show that post-processing improves the quality of decoded images for human in scalable image coding methods.

\begin{figure}[t]
  \centering
  \includegraphics[width=0.48\textwidth]{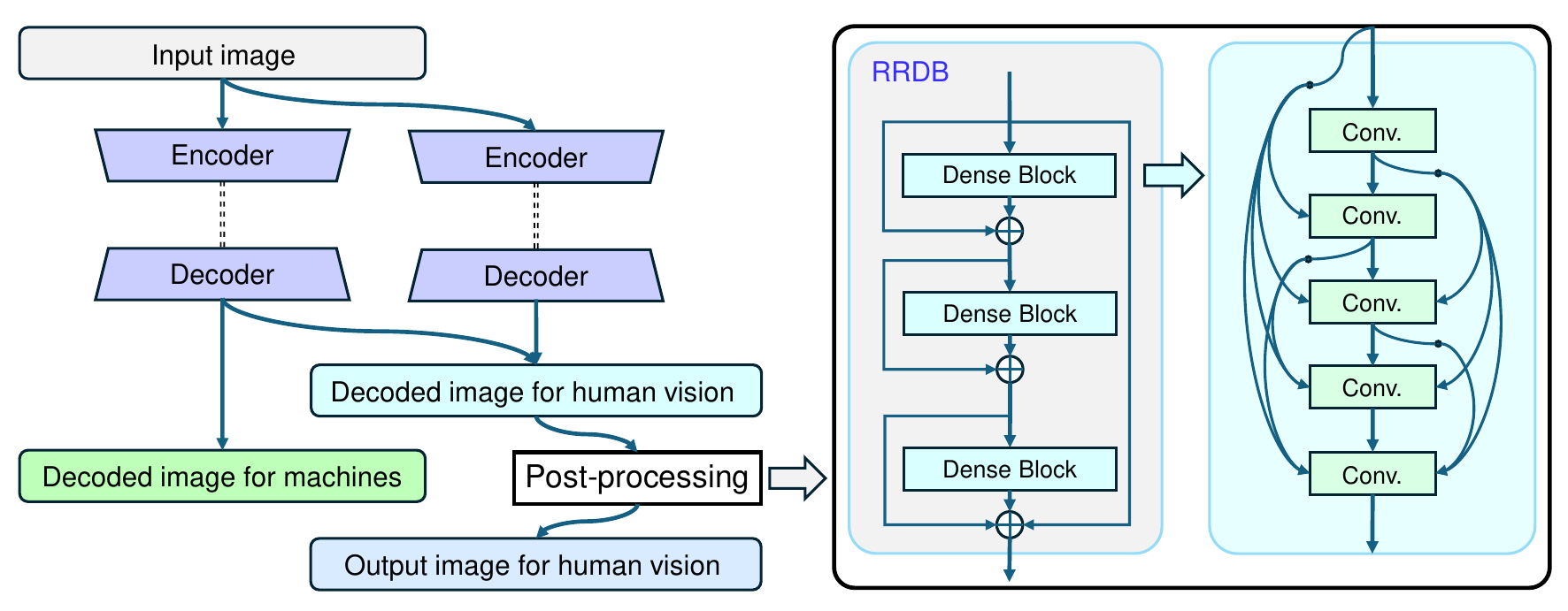}
  \caption{Image compression process and structure of the post-processing model.}
  \label{proc}
\end{figure}

\section{Related Works}
\subsection{Scalable Image Coding for Humans and Machines}
There are ongoing research in scalable image coding methods that cater to both image recognition models and human vision\cite{a5,a6}.
This field is known as "scalable image coding for humans and machines," and several methods have been proposed.
These methods decode images for humans by adding supplementary information to the images decoded for machines.
Learned Image Compression (LIC)\cite{a7}, a neural network-based image compression technique, is frequently used in this area.
By combining the LIC model for machines with the LIC model for additional information, a method for decoding images suitable for both machines and humans has been developed\cite{a5}.
The inclusion of SA-ICM\cite{a8} in scalable coding method enables image compression for various image recognition models and humans.
SA-ICM is one of the image coding methods for machines and is not optimized for a specific image recognition model\cite{a9}.
Hence, scalable image coding methods incorporating with SA-ICM can compress images for various image recognition models and humans.

\subsection{Post-processing for Decoded Images}
Post-processing for decoded images is an effective method for enhancing both image quality\cite{a3} and image recognition accuracy\cite{a4}.
Numerous neural network-based models for post-processing have been proposed to improve compression performance.
These models are trained to make the encoded image more closely resemble the original image.
The Mean Squared Error (MSE) between the original and output images is often used as the loss function.

\section{Proposed Method}
We propose a method to enhance the image quality of compressed images by incorporating a post-processing model into the scalable coding method.
As a scalable coding method, we employ a method that progressively compresses images for both humans and machines by linking two LIC models\cite{a5}.
The decoded image produced by this method exhibits specific coding noise compared to the LIC model typically used for human decoding.
To address this noise, we aim to enhance image quality by training the post-processing model to learn and mitigate this unique noise. 
We adopt the Residual in Residual Dense Block (RRDB) structure proposed by X.Wang \textit{et al.}, which is widely utilized for tasks like image super-resolution and denoising\cite{a10}.
Fig. \ref{proc} illustrates the model structure and its position in the image compression pipeline. 
\begin{figure}[t]
  \centering
  \includegraphics[width=0.48\textwidth]{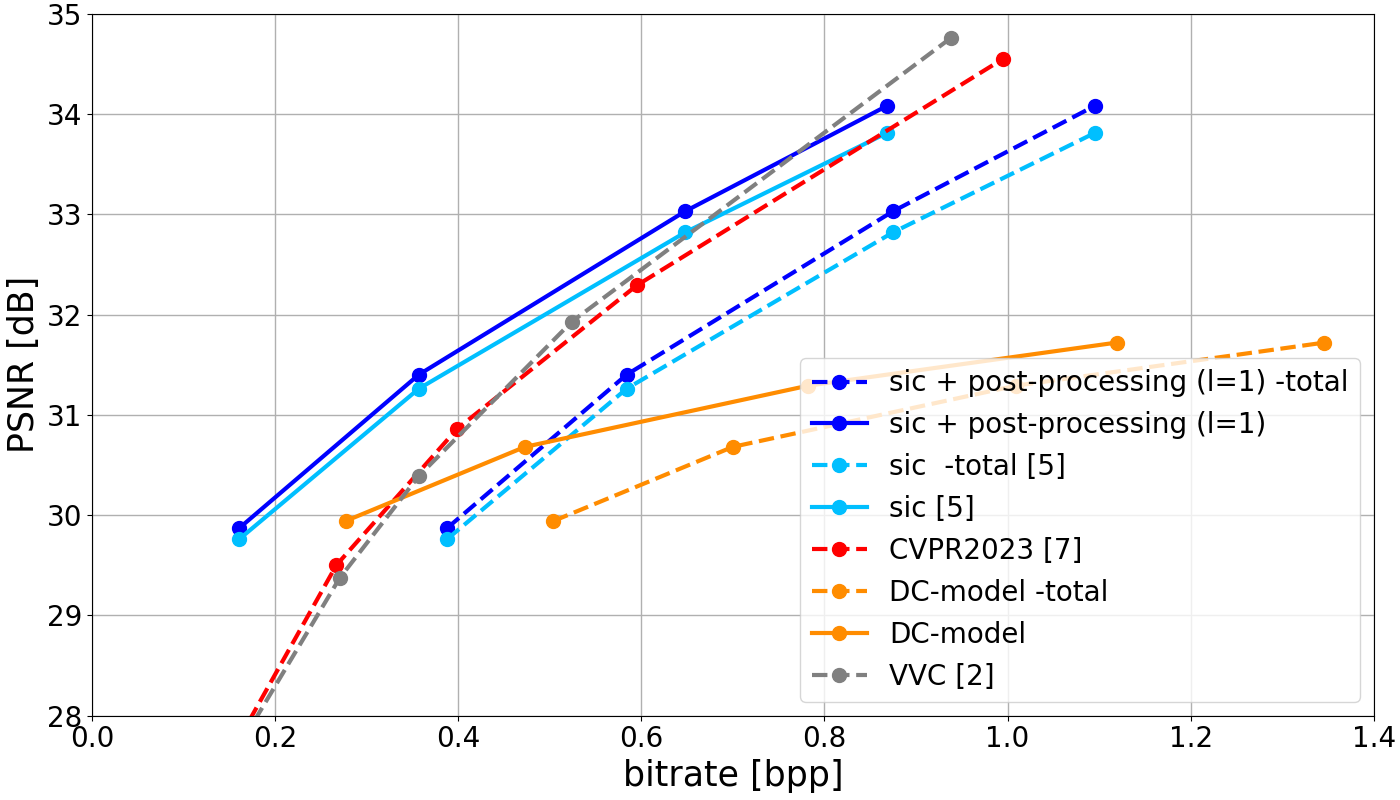}
  \caption{Compression performance of the proposed and comparative methods. "sic” stands for scalable image coding.}
  \label{fig:psnr}
  \end{figure}

\begin{table}[t]
  \centering
  \caption{Image quality (PSNR) in the decoded image} \label{tab:psnr}
  \small
  \begin{tabular*}{8.8cm}{c|cccc}
      \hline
      \multirow{2}{*}{method} & \multicolumn{4}{c}{$\lambda$} \\
                               & 0.005 & 0.010 & 0.020 & 0.030 \\
      \hline
      \hline
       w/o post-processing         & 29.76 & 31.26 & 32.82 & 33.81\\
       w/ post-processing ($l=1$)  & 29.87 & 31.40 & 33.03 & 34.08\\
       w/ post-processing ($l=2$)  & 29.88 & 31.42 & 33.05 & 34.09\\
      \hline
  \end{tabular*}
  \end{table}

\section{Experiment}
We utilize the COCO dataset\cite{a11} to train and evaluate post-processing model.
Initially, we compress all images in the dataset for human vision, using a scalable coding method\cite{a5}.
Subsequently, using the COCO-Train dataset prepared in this manner, we train the post-processing model using MSE-loss function.
To compare experimental results across post-processing model sizes, we train one model with a single RRDB and another with two RRDBs.
Once trained, the models are applied to the compressed images in the COCO-Val dataset to mitigate encoding noise.

\begin{figure}[t]
  \centering
  \includegraphics[width=0.48\textwidth]{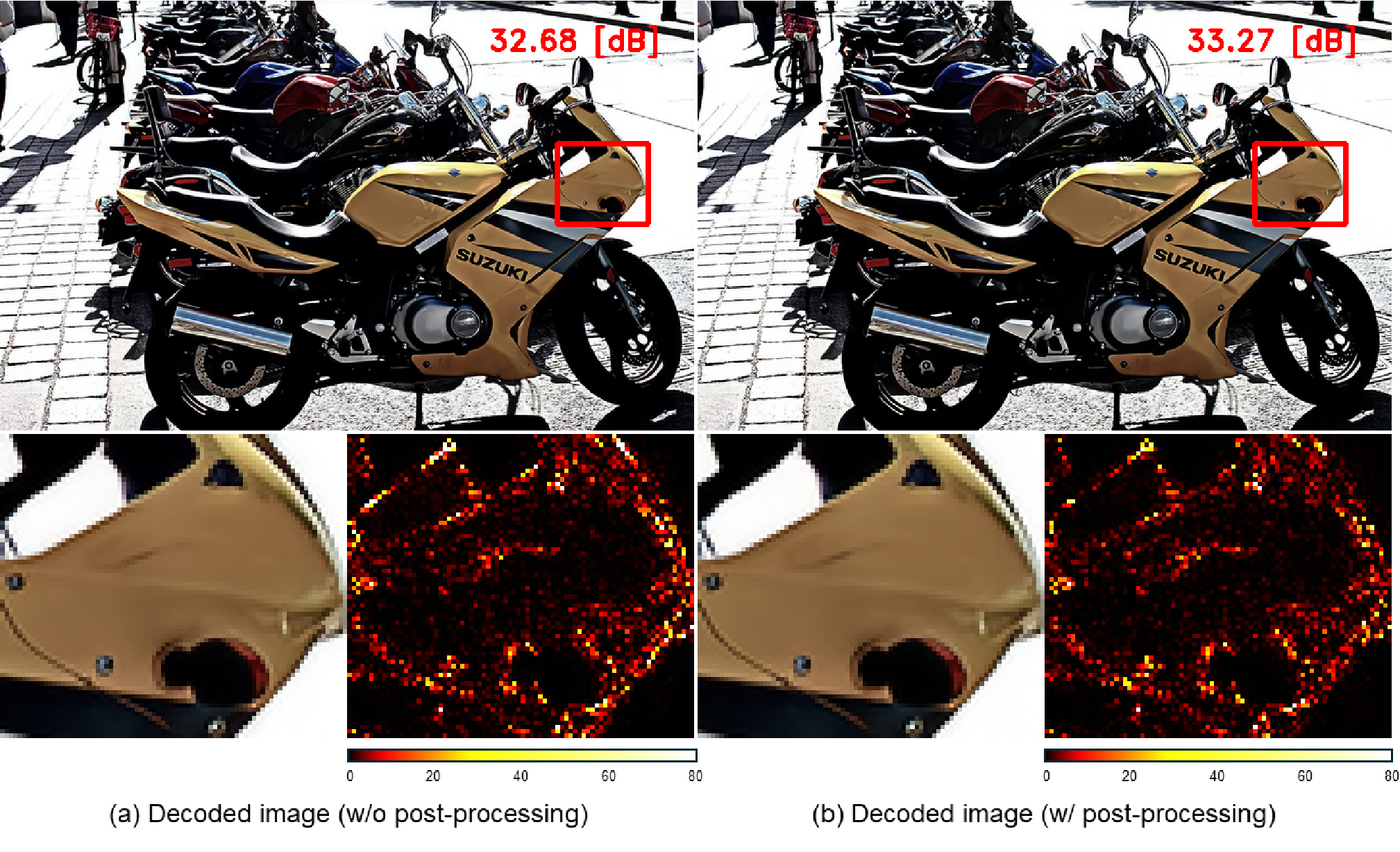}
  \caption{An example of the decoded images with and without post-processing and the coding noise in those images. Areas with large coding noise are shown brighter.}
  \label{fig:ex}
\end{figure}

The image quality of with and without post-processing are depicted in Table \ref{tab:psnr} and Fig.\ref{fig:psnr}.
In Table \ref{tab:psnr}, the value of $l$ represents the number of RRDBs in the model structure and the $\lambda$ value is a variable to control the bitrate.
In Fig.\ref{fig:psnr}, a comparison of the image compression performance between the proposed and conventional methods is presented.
The performance of our proposed method is represented by the blue curve, while the performance of the original scalable image coding method is depicted by the light blue curve.
The dotted line illustrates the total amount of additional information and image information for machines in scalable image coding. 
The solid line shows only the amount of additional information.
Additionally, Fig.\ref{fig:ex} presents examples of the input and output images from the post-processing model, as well as the noise present in each image. 
These findings show the effectiveness of the post-processing in scalable image coding.
\section{Conclusion}
In this paper, we propose a post-processinging method aimed at eliminating coding noise in scalable image coding method.
Experimental results demonstrate that employing the post-processing model enhances the image quality of the decoded image.
Future research endeavors should involve the incorporation of evaluation metrics like SSIM and LPIPS, as well as an exploration of methods to enhance image quality.

\section*{Acknowledgment}
The results of this research were obtained from the commissioned research (JPJ012368C05101) by National Institute of Information and Communications Technology (NICT), Japan.

\end{document}